\documentclass[10pt,twocolumn,letterpaper]{article}

\usepackage{cvpr}
\usepackage{times}
\usepackage{epsfig}
\usepackage{graphicx}
\usepackage{amsmath}
\usepackage{amssymb}
\usepackage{color}
\usepackage{slashbox}
\usepackage[flushleft]{threeparttable}

\usepackage[colorlinks=true,linkcolor=black,citecolor=black,urlcolor=black,breaklinks=true,bookmarks=false]{hyperref}

\cvprfinalcopy 

\def\cvprPaperID{****} 

\begin{document}

\title{ Vehicle's Lane-changing Behavior Detection}

\author{
Mengwen He\\
CyLab-ECE\\
4720 Forbes Ave. CIC 2223G\\
{\tt\small mengwenh@andrew.cmu.edu}
\and
Iljoo Baek\\
CyLab-ECE\\
4720 Forbes Ave. CIC 2224G\\
{\tt\small ibaek@andrew.cmu.edu}
}

\maketitle

\begin{abstract}
   The lane-level localization accuracy is very important for autonomous vehicles. The Global Navigation Satellite System (GNSS), e.g. GPS, is a generic localization method for vehicles, but is vulnerable to the multi-path interference in the urban environment. Integrating the vision-based relative localization result and a digital map with the GNSS is a common and cheap way to increase the global localization accuracy and thus to realize the lane-level localization. This project is to develop a mono-camera based lane-changing behavior detection and tracking algorithm module for the correction of lateral GPS localization. We implemented a Support Vector Machine (SVM) based framework to directly classify the driving behavior, including the lane keeping, left and right lane changing, from a sampled data of the raw image captured by the mono-camera installed behind the window shield. The training data was collected from the driving around Carnegie Mellon University, and we compared the trained SVM models w/ and w/o the Principle Component Analysis (PCA) dimension reduction technique. Next, we intend to compare the SVM based classification method with the CNN method.
\end{abstract}

\section{Introduction}

The autonomous vehicle highly relies on the accurate localization technique because it enables reliable planning and control operations for the safe autonomous driving. The Global Navigation Satellite System (GNSS), e.g. GPS, GLONASS, Beidou (Compass), and Galileo, provides commercial localization devices with affordable price but low accuracy to vehicles. It works well while driving on the highway; however, it is vulnerable to the multi-path interference caused by trees, buildings, or overhead bridges in the urban area.

The prevalent method to enhance the global localization accuracy is to use 3D Light Detection and Ranging (LiDAR) sensors, e.g. Velodyne, to conduct registration with 3D point-cloud map. This can guarantee centimeter level localization accuracy \cite{takeuchi20063} so that this method can be found on various commercial or experimental autonomous vehicles like Google, Baidu, Uber, and Toyota. However, this method is very expensive because of the 3D LiDAR and dense point-cloud map.

Therefore, the common and cheaper camera-based methods are preferred for affordable autonomous vehicles. The visual odometry method can achieve decimeter level relative localization \cite{merfels2016pose}. By implementing filter-based, e.g. Kalman filter and its variants, or graph-based, i.e. non-linear least squares, methods, we can fuse the GNSS global localization with the visual odometry relative localization to get an enhanced global result.

\begin{figure}[t]
	\centering
	\includegraphics[width=0.4\textwidth]{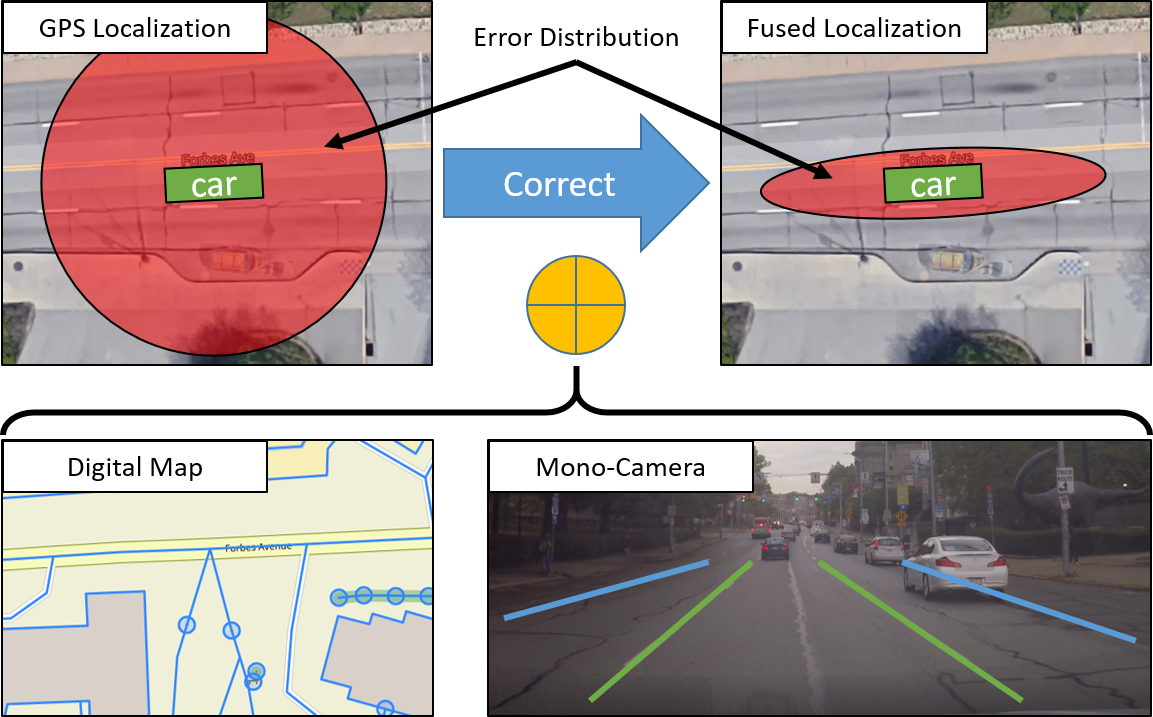}
	\caption{The illustration of our project's objective. We get a global localization result from GPS with a Gaussian error model. By integrating the digital map and the lane information from our lane-changing behavior detection and tracking algorithm, we can laterally correct the global localization error.}
	\label{fig:intro}
\end{figure}

In this project, we want to realize a vehicle's lane-changing behavior detection and tracking algorithm based on a mono-camera. Following the integration of GNSS and camera, we want to use the road lane information from a mono-camera to enhance the global localization result; therefore, the detection and tracking of the lane-changing behavior is required to tell which lane the vehicle is on. Coupled with a digital map with road lane information, we can laterally reduce the global localization error from the GNSS as shown in Fig.\ref{fig:intro}.

\begin{figure}[t]
	\centering
	\includegraphics[width=0.4\textwidth]{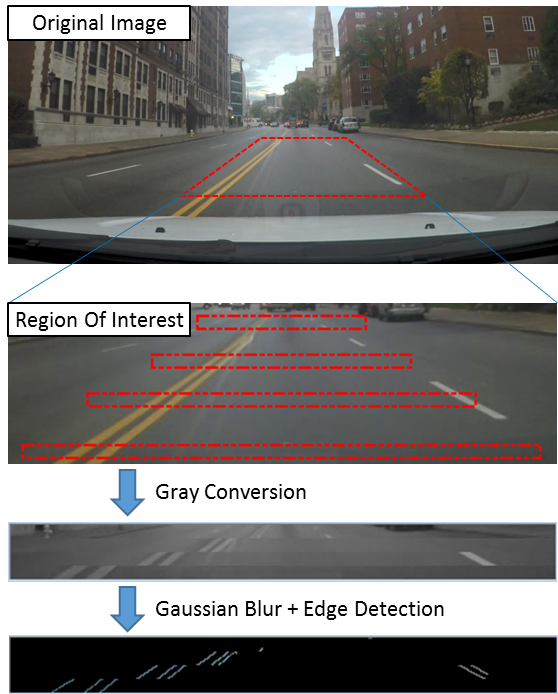}
	\caption{Top: the original image with a ROI in front of the vehicle. Middle: The ROI has four sample layers (red rectangles) and each layer represents a distance (1m, 10m, 20m, 30m) from the vehicle. Bottom: Stack the sampled layers of processed pixels (edge extraction on gray image) row by row to form the feature vector.}
	\label{fig:sample}
\end{figure}

We employed a Support Vector Machine (SVM) based framework to detect the lane-changing behavior, and the training and test data/feature is directly sampled from the raw image's region-of-interest (ROI) as shown in Fig.\ref{fig:sample}. Therefore, we performed a Principle Component Analysis (PCA) dimension reduction technique to compress the feature's dimensionality as well as to keep more than $98\%$ of its original energy. 

Morevoer, because of the limitation of data size, we choose to use SVM instead of the Convolution Neural Network (CNN) \footnote{18-794 Pattern Recognition Theory homework assignment 3} to classify the lane-changing behavior as lane-keeping, right lane-changing, or left lane-changing. But, we plan to compare the training results between the SVM and CNN using the same training data.

We conducted the experiment around Carnegie Mellon University with a Go-Pro HD Camera (Hero 4, $1920 \times 1080$) mounted after the window shield of Iljoo's vehicle. We used the off-the-shelf LibSVM \cite{chang2011libsvm} to train the lane-changing behavior classfier w/ or w/o PCA dimension reduction.

\section{Related Work}

We first referred to \cite{mandalia2005using} which uses the SVM for lane-change detection. The features they used are all from the vehicle's CANBUS containing the steering angle, speed, etc. However, becasue this method does not use the camera or other external sensors to observe the environment, especially the strongly related lanes, it cannot perform perfectly for the real application, e.g. on the curved road or under the driving situation with large speed variance. \cite{pentland1999modeling,kuge2000driver} both implemented Dynamic/Hidden Markov Models to characterize and detect driving maneuvers. They used a simulation platform to collect drivers' behavior as well as the simulated vehicle's steering angle. However, this method only provides the simulated experiment results, and it is necessary to develop more general models and assure robustness corresponding to actual driving situations. \cite{olsen2003modeling} conducted many real experiments with cameras which observe the front as well as the driver's appearance and behavior. This dissertation's objectives are 1) to characterize normal lane changes in which a slow lead vehicle was present, 2) develop predictive models distinguishing baseline (straight-ahead) driving from lane changes, and 3) provide design guidelines relevant to the issuing of lane change collision warnings. After statistically analyzing the collected huge dataset, the author implemented a Logistic Regression to build the predictive model for lane-change behavior. The result is good, however, in this project, we don't have too much time or such resources to follow his work. In \cite{salvucci2004inferring}, the author try to map a driver's observable actions to their unobservable intentions, say lane-change, and this method is named as mind-tracking. To build this map, they implemented many cognitive models of driving behaviors to inference the driver's behavior. However, similar to \cite{mandalia2005using,pentland1999modeling,kuge2000driver}, we think this method still does not touch the direct clue for lane-change detection, the relative position between ego-vehicle and lanes captured by camera.

\section{Approach and Algorithm}

\subsection{Assumptions}

For the time limitation, we make some assumptions to simplify this project; however, in the future research, we will gradually get rid of these assumptions.

\begin{itemize}
	\item This project will only work on the road with clear lane markers, and the intersection is not in its scope.
	\item The occlusion caused by other vehicles will not be considered in our training and test data.
	\item An accurate global digit map already exists and contains the lane information.
	\item The initially occupied lane is known from an upstream algorithm module.
\end{itemize}

\subsection{Training Data Collection and Labeling}

\begin{figure}[t]
	\centering
	\includegraphics[width=0.3\textwidth]{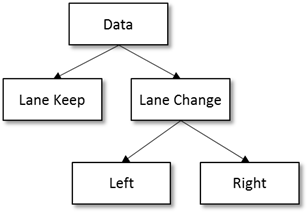}
	\caption{The two layer decision tree converts the three-class classification problem into a two-step binary classification problem.}
	\label{fig:label}
\end{figure}

While training the SVM, it is very important to choose the most effective set of features that can represent significant differences during a lane change against lane keeping. We have tried three types of methods to collect the desired features in thie project. For the labeling, the classifier has two layers of decision as shown in Fig.\ref{fig:label}. It firstly figures out whether the vehicle changes the lane or not, and then it finds the direction of lane change.

\subsubsection{Hough Transformation Based Method}
\begin{figure}[t]
	\centering
	\includegraphics[width=0.45\textwidth]{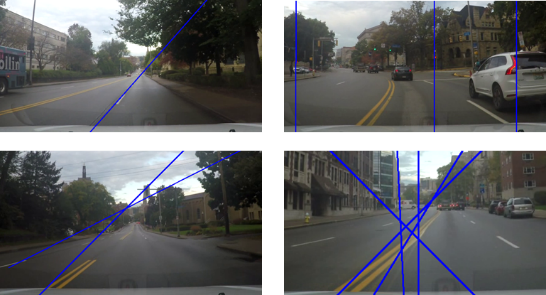}
	\caption{Top left/right \cite{lanedetection}:  Basic Hough transformation was conducted with canny edge detection result to find and fit lines representing the lane markers. It failed to detect the lanes in curved road scenes. Bottom left/right \cite{carcv}: Color segmentation and background subtraction were applied to extract lane marker features. Then the features were fitted into lines using the Hough transformation and contour estimation.}
	\label{fig:lanedetection}
\end{figure}

We firstly tried to extract the lane position information by detecting lanes using open source lane detectors \cite{lanedetection,carcv}, and both of them implement the Hough transformation in the OpenCV to extract the lane edge. However, the detectors could not provide robust feature information because the ratio of true negative was too high. Additionally, the frequent false positive result (e.g. Fig.\ref{fig:lanedetection}) from the detectors has a negative impact on the SVM's performance.

\subsubsection{Deep Neural Network Based Method}
\begin{figure}[t]
	\centering
	\includegraphics[width=0.4\textwidth]{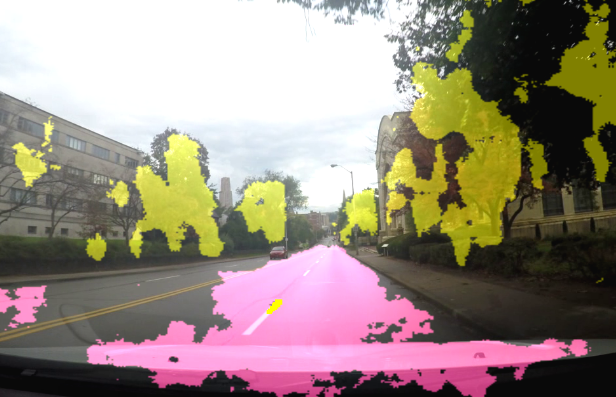}
	\caption{Each pixel in the image is classified through a sequence of non-linear processing layers (encoders) and a corresponding set of decoders followed.}
	\label{fig:segnet}
\end{figure}

Then we used the deep learning based SegNet tool \cite{segnet, badrinarayanan2015segnet2} to distinguish the lane markers from a road scene. However, the lane markers were often misclassified as the road in the segmentation result as shown in Fig.\ref{fig:segnet}. This is also not suitable for the SVM based classifier.

\subsubsection{ROI Sampled Edge Features}

We tried to extract the binary edge information from a pre-defined ROI as shown in Fig.\ref{fig:sample}. The ROI has four layers and each layer represents a distance (1m, 10m, 20m, 30m) from the vehicle. The ROI pixels are converted to gray, blurred by Gaussian, and filtered by Canny Edge detector. According to our test, these binary edge features are not stable enough to provide continuous information for the SVM classifier.

\subsubsection{ROI Sampled HOG descriptor}

\begin{figure}[t]
	\centering
	\includegraphics[width=0.4\textwidth]{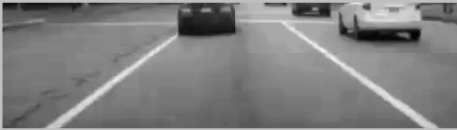}\\
	\includegraphics[width=0.4\textwidth]{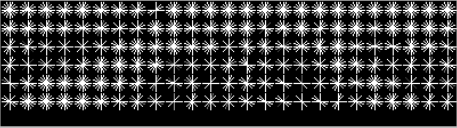}
	\caption{The HoG gradients in 16x16 cell with 8 orientations provides robust features to distinguish lane markers and their orientations.}
	\label{fig:hog}
\end{figure}

Lastly, The Histograms of Oriented Gradients(HoG)\cite{dalal2005histograms} was applied to a rectangular area which includes the four layers and intermediate areas between the layers as shown in Fig.\ref{fig:hog}. We converted the sub image to gray, and computed gradients in 16x16 cells with 8 orientations. We did not conduct block overlap normalization because our test result showed insignificant performance improvement with the block normalization. 

\subsection{PCA Dimension Reduction}

The training data/feature's dimensionality is 16,000; therefore, we want to use the PCA to shrink its dimensionality and focus on the main components. Moreover, because the dimensionality is greatly larger than the data size, we will use the Gram Matrix Trick to accelerate the eigen decomposition operation.

\subsubsection{Gram Matrix Trick}

Given a centralized training data matrix $X$ with dimension $d \times N$, we derive its covariance matrix as $\Sigma=E(XX^{T})$. The PCA needs to solve the problem as below:
\begin{equation}
\begin{array}{rcl}
\Sigma \vec{v} & = & \lambda \vec{v} \\
XX^T\vec{v} & = & \lambda' \vec{v} \\
X^TXX^T\vec{v} & = & \lambda' X^T \vec{v} \\
X^TX\vec{v}' & = & \lambda'' \vec{v}'~~~~(\vec{v}'=\eta X^T\vec{v}) \\
\end{array}
\end{equation}
Therefore, we only need to solve the eigen decomposition of the Gram matrix $X^TX$ with dimension $N \times N$. To get the final eigenvectors $\{\vec{v}_i\}$, we need to solve:
\begin{equation}
\left\{\begin{array}{rcl}
XX^T\vec{v} & = & \lambda' \vec{v} \\
\vec{v}' & = & \eta X^T \vec{v} \\
\end{array}\right. \Rightarrow \vec{v}=\eta' X\vec{v}'
\end{equation}
Then we can use $\Sigma\vec{v}=\lambda\vec{v}$ to derive all the corresponding eigenvalues $\{\lambda_i\}$.

\subsubsection{Dimension Reduction}

Firstly, We sort the derived eigenvalues (associated with the corresponding eigenvectors) in a descending order, and we calculate the total energy as below:
\begin{equation}
\Lambda=\sum_{i=1}^{d}{\lambda_i}
\end{equation}
Then, we choose the first $M$ biggest eigenvalues whose summation is just larger than a threshold ratio $r$ of the total energy $\Lambda$.
\begin{equation}
\begin{array}{rcl}
\sum_{i=1}^{M}{\lambda_i} & \geq & r\Lambda \\
\sum_{i=1}^{M-1}{\lambda_i} & < & r\Lambda \\
\end{array}
\end{equation}
Finally, we use the first $M$ biggest eigenvalues' corresponding eigenvectors to form a PCA dimension reduction matrix $P$ as below:
\begin{equation}
P=[\vec{v}_1,\dots,\vec{v}_M]
\end{equation}
Therefore, the dimension reduced new centralized trainidng data matrix is $X'=P^TX$. If we use the PCA dimension reduction on the training data as well as the SVM, we also need to apply the dimension reduction on the test data $Y$ following these two steps:
\begin{enumerate}
	\item Centralize the test data with the mean ($\mu_X$) of the training data.
	\item Reduce its dimension to get $Y'=P^T(Y-\mu_X)$
\end{enumerate}

\subsection{LibSVM Training and Test}

The LibSVM already provides some off-the-shelf applications to conduct and evaluate the binary classification, and what we need to provide is the training and test data following the format required by the LibSVM. The data is stored in ASCII format, and each row in it presents a labeled feature. Because the LibSVM supports sparse matrix operations, the labeled feature is stored as a label followed with a sequence of indexed non-zero values as below:

\begin{verbatim}
-1 1:1 11:1 18:1 20:1 37:1 42:1 59:1 
+1 5:1 18:1 19:1 39:1 40:1 63:1 
\end{verbatim}

\section{Experiments and Progress}

\begin{figure}[t]
	\centering
	\includegraphics[width=0.4\textwidth]{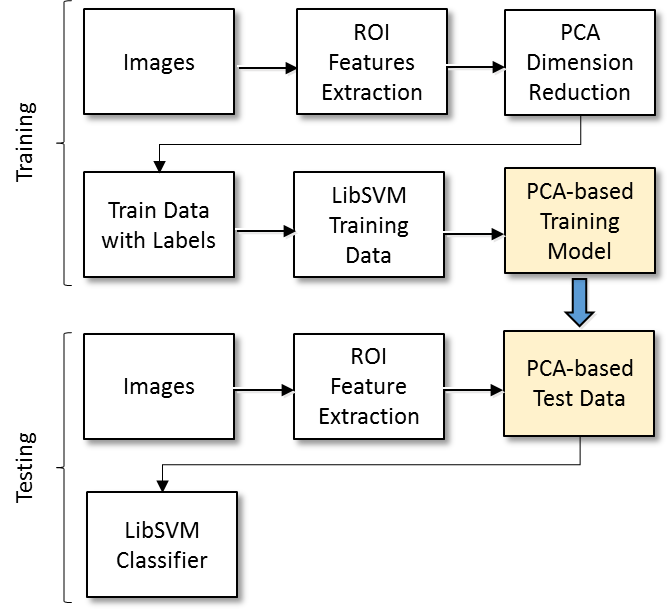}
	\caption{The pipeline of this SVM based lane-changing behavior detection project.}
	\label{fig:pipeline}
\end{figure}

The implementation pipeline of this project is shown as Fig.\ref{fig:pipeline}.

\subsection{Data Collection and Labeling}

\begin{figure}[t]
	\centering
	\includegraphics[width=0.5\textwidth]{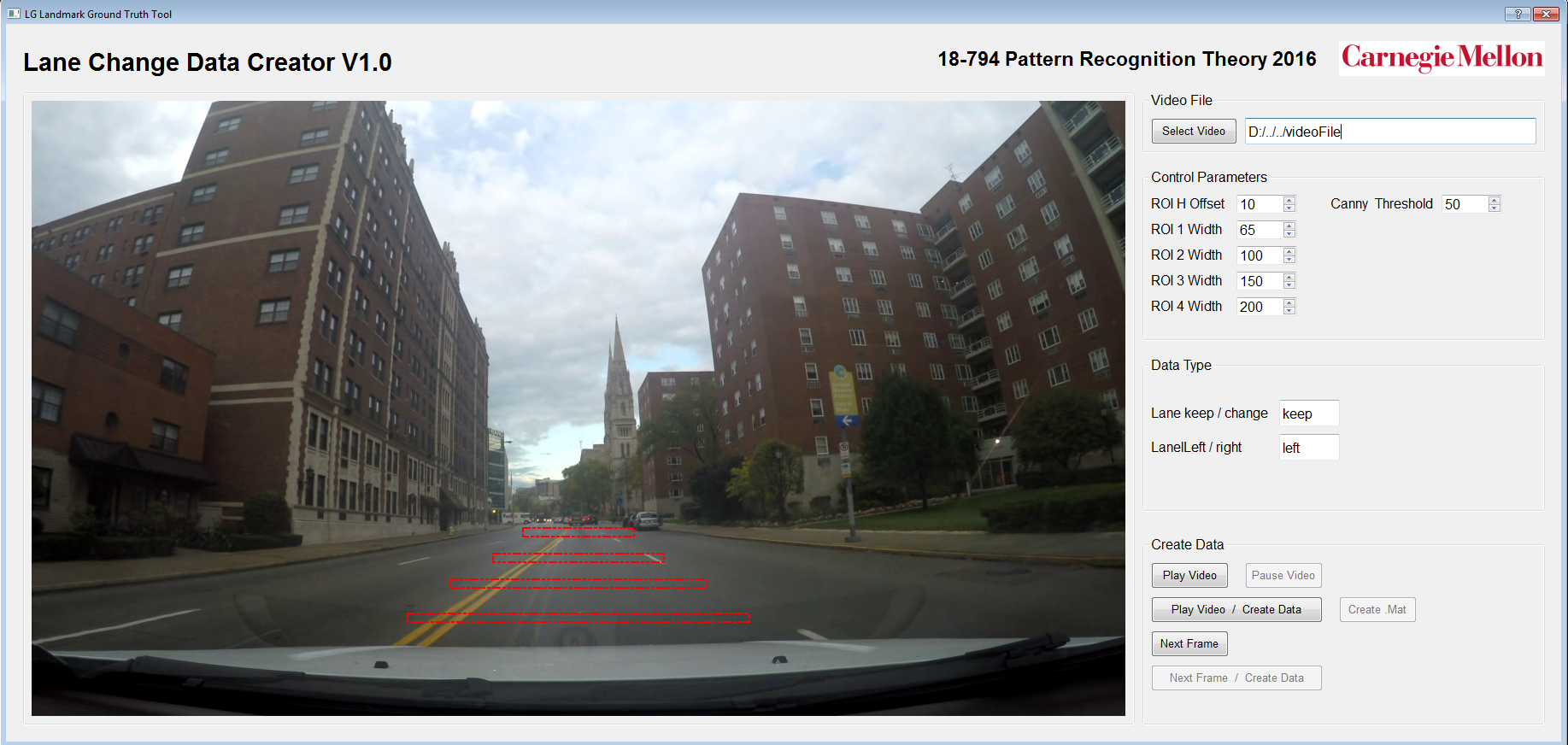}
	\caption{The tool to collect data and extract feature.}
	\label{fig:tool}
\end{figure}

\begin{figure}[t]
	\centering
	\includegraphics[width=0.4\textwidth]{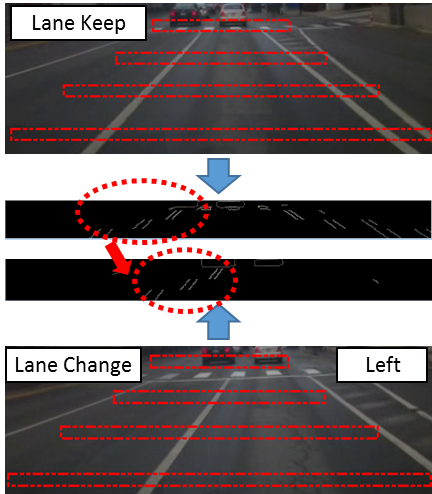}
	\caption{Compare the Canny features extracted from the `Lane Keep' behavior and the `Lane Change - Left' behavior.}
	\label{fig:feature}
\end{figure}

\begin{figure}[t]
	\centering
	\includegraphics[width=0.5\textwidth]{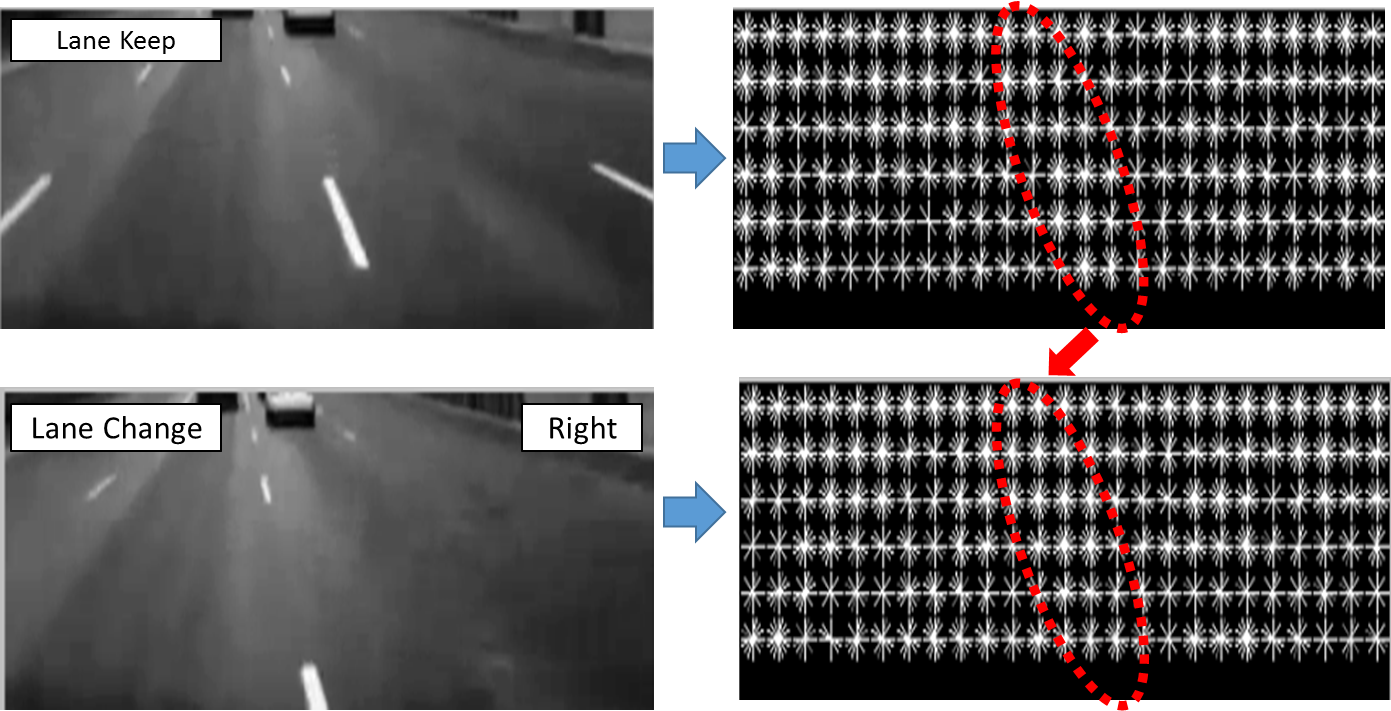}
	\caption{Compare the HoG features extracted from the `Lane Keep' behavior and the `Lane Change - Righteft' behavior.}
	\label{fig:hogfeature}
\end{figure}

A tool shown in Fig.\ref{fig:tool} was developed to create the training and ground truth data. We defined four classes which are `Lane Keep', `Lane Change - Left', `Lane Change- Right', and `Unknown' (when the vehicle is at intersections where no lane information is available). The class type is manually determined, and the feature data is automatically labeled using the user interface provided by this tool. The binary Canny edge feature presents the distinguishing pattern between two class types as shown in Fig.\ref{fig:feature}, and the difference of HoG features between two class types is shown in Fig.\ref{fig:hogfeature}.

\subsection{SVM Training Result}

\begin{figure}[t]
	\centering
	\includegraphics[width=0.4\textwidth]{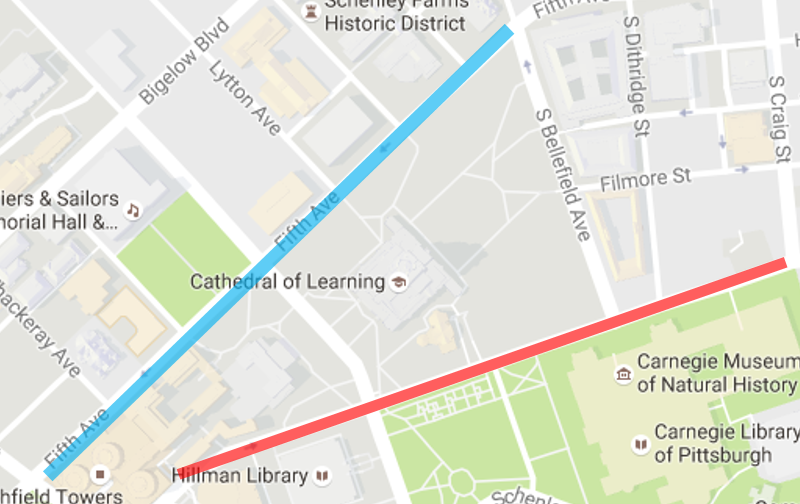}
	\caption{The locations where we collect the training data (blue) and test data (red).}
	\label{fig:datamap}
\end{figure}

The training and test data are collected from the roads around Carnegie Mellon University as shown in Fig.\ref{fig:datamap}.  From these two roads, we extracted and labeled $4059$ training data and $1611$ test data.

We compared the accuracy rate between using Canny features and HoG features as shown in Tab.\ref{tab:accuracy}. For the canny feature, we also compared the classification results w/ and w/o PCA dimension reduction pre-processing. The performance gap between linear SVM and Kernel SVM was narrow in our test because most of the features in the ROI are regularly distributed around left and right lane markers. But the performance of HoG feature is better than that of Canny feature. The canny feature can extract the edge of lanes (also work on other object like vehicle, curb, etc.), but it only represents the edge in pixel level. However, the HoG feature can not only response to the lane's edge but also represent its orientation. Therefore, the HoG feature is more suitable than the canny feature for the lane-change behavior detection.

\begin{table}[!h]
	\centering
	\caption{SVM Based Classifiers' Accuracy Rate}
	\label{tab:accuracy}
	\begin{tabular}{|c|c|c|c|}
		\hline
		Feature	&	Pre-Process	&	L-SVM	&	K-SVM	\\
		\hline
		Canny	&	w/o PCA (16,000) & 53.4\% & 55.9\% \\
				&	w/ PCA (362)	 & 53.1\% & 55.9\% \\
		\hline
		HoG		&	N/A (1200)	& 66.9\% & 68.5\% \\
		\hline
	\end{tabular}
\end{table}

\begin{figure}
	\centering
	\includegraphics[width=0.47\textwidth]{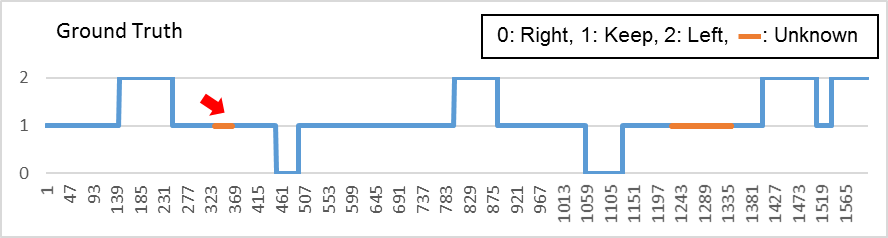}\\
	\includegraphics[width=0.47\textwidth]{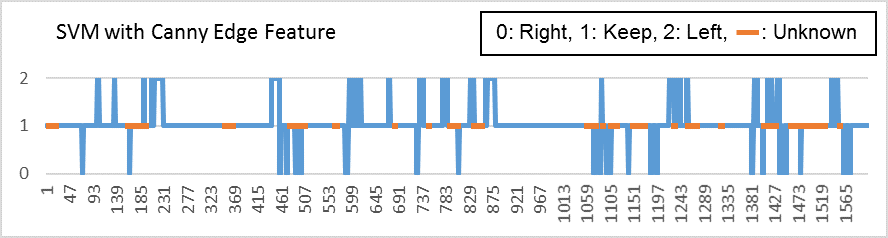}\\
	\includegraphics[width=0.47\textwidth]{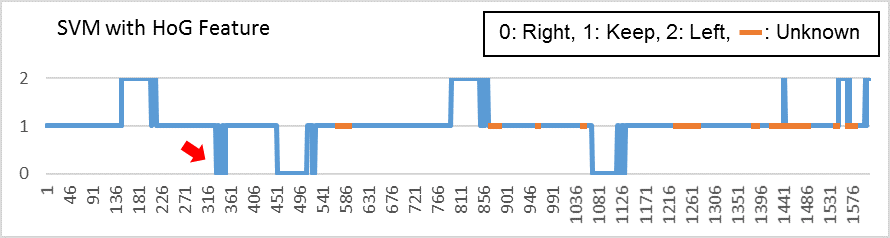}\\
	\includegraphics[width=0.47\textwidth]{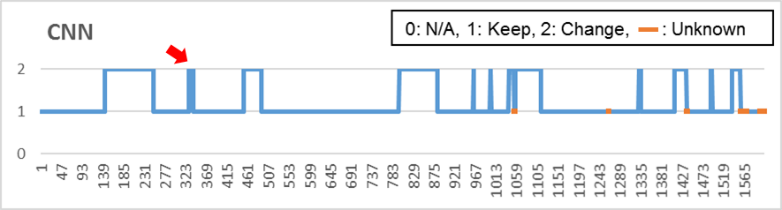}\\
	\caption{The detailed classification results using SVM-Canny, SVM-HoG, and CNN. The red arrow indicates a mislabeling test data segment, which should be right lane-change.}
	\label{fig:show}
\end{figure}

Additionally, we use the confusion matrices to show the details of each SVM based classifier's accuracy rate as shown in Tab.\ref{tab:cm1},\ref{tab:cm2},\ref{tab:cm3}. The SVM classifier using Canny edge feature completely failed to detect lane-change behavior due to the insufficiency of edges in the test scenes and so many noisy features from the front obstacles in both the training and testing data sets. Because of the orientation representation in HoG feature, the HoG based SVM classifier can efficiently detect the lane-change behavior. However, it generated so many miss classifications between `Keep' and `Unknown'. But in this test scene, the vehicle really kept its direction (a kind of `Keep') during the `Unknown' data segments, and at the most time of approaching intersections, the ROI contains the lane information of the road after the intersection. Therefore, if we omit these miss classifications, the accuracy rate can reach 80\%. To further investigate the performance of the classifiers, we present the classification results compared to the ground truth as shown in Fig.\ref{fig:show}.

\begin{table}[!h]
	\centering
	\caption{Confusion Matrix using Canny Features}
	\label{tab:cm1}
	\begin{tabular}{|c|c|c|c|c|}
		\hline
		\backslashbox{GT}{Res.}		&	Keep	&	Left	&	Right	&	Unknown	\\
		\hline
		Keep	&		900	&		0	&		0	&	0		\\
		\hline
		Left	&		305	&		0	&		0	&	0		\\
		\hline
		Right	&		124	&		0	&		0	&	0		\\
		\hline
		Unknown	&		281	&		0	&		0	&	0		\\
		\hline
	\end{tabular}
\end{table}

\begin{table}[!h]
	\centering
	\caption{Confusion Matrix using Hog Features and L-SVM}
	\label{tab:cm2}
	\begin{threeparttable}
	\begin{tabular}{|c|c|c|c|c|}
		\hline
		\backslashbox{GT}{Res.}		&	Keep	&	Left	&	Right	&	Unknown	\\
		\hline
		Keep	&	808  &  22  &  21  &  49	\\
		\hline
		Left	&	135  & 133  &   0  &  37	\\
		\hline
		Right	&	23   &  0   &  94  &   7	\\
		\hline
		Unknown	&	214* &  5   &  19  &  43	\\
		\hline
	\end{tabular}
	\begin{tablenotes}
		\footnotesize
		\item[*] Classification accuracy can be increased to 80\% if this unknown misclassification is considered as true positive.
	\end{tablenotes}
	\end{threeparttable}
\end{table}

\begin{table}[!h]
	\centering
	\caption{Confusion Matrix using HoG Features and K-SVM}
	\label{tab:cm3}
	\begin{threeparttable}
	\begin{tabular}{|c|c|c|c|c|}
		\hline
		\backslashbox{GT}{Res.}		&	Keep	&	Left	&	Right	&	Unknown	\\
		\hline
		Keep	&	826  &  19  &  24  &  31	\\
		\hline
		Left	&	129  & 135  &   0  &  41	\\
		\hline
		Right	&	28  &   0   &  96  &   0	\\
		\hline
		Unknown	&	215*  &   5  &  15  &   46	\\
		\hline
	\end{tabular}
	\begin{tablenotes}
		\footnotesize
		\item[*] Classification accuracy can be increased to 80\% if this unknown misclassification is considered as true positive.
	\end{tablenotes}
	\end{threeparttable}
\end{table}

\begin{table}[!h]
	\centering
	\caption{Confusion Matrix using CNNs}
	\label{tab:cm4}
	\begin{tabular}{|c|c|c|c|}
		\hline
		\backslashbox{GT}{Res.}		&	Keep	&	Change (L\&R)	&	Unknown	\\
		\hline
		Keep			&		951	&		22	&		3	\\
		\hline
		Change (L\&R)	&		126	&		331	&		28	\\
		\hline
		Unknown			&		134	&		14	&		2	\\
		\hline
	\end{tabular}
\end{table}

\subsection{CNN Training Result}

To measure the efficiency of our SVM based lane-change behavior detector, we implemented a CNN based classifier as a baseline. The input of this network is the original image in the same ROI defined in Fig.\ref{fig:sample}, and the network has 10 layers as Tab.\ref{tab:cnn}. We only trained this network to detect the lane-change behavior (the first layer in Fig.\ref{fig:label}); therefore for CNN test, we only have three labels: `Keep', `Change', and `Unknown'. The training accuracy is $99.93\%$, and the test accuracy rate is $79.70\%$. The confusion matrix of using CNN is shown in Tab.\ref{tab:cm4}, and the detailed performance is shown in Fig.\ref{fig:show}.

\begin{table}[!h]
	\centering
	\caption{The Construction of CNNs}
	\label{tab:cnn}
	\begin{tabular}{|c|c|c|c|}
		\hline
		\#		&	Type		&	Dimension						&	Stride	\\
		\hline
		0		&	Input		& 	$110\times400\times3\times1$	&	1		\\
		\hline
		1		&	Conv		&	$5\times5\times3\times40$		&	1		\\
		\hline
		2		&	Pool (Max)	&	$2\times4$						&	$2\times4$\\
		\hline
		3		&	Conv		&	$5\times5\times40\times80$		&	1		\\
		\hline
		4		&	Pool (Max)	&	$2\times4$						&	$2\times4$\\
		\hline
		5		&	Conv		&	$5\times5\times80\times160$		&	1		\\
		\hline
		6		&	Pool (Max)	&	$2\times2$						&	$2\times2$\\
		\hline
		7		&	Conv		&	$5\times5\times160\times1600$		&	1	\\
		\hline
		8		&	ReLU		&	N/A								&	N/A		\\
		\hline
		9		&	Conv		&	$6\times5\times1600\times3$		&	1		\\
		\hline
		10		&	SoftMaxLoss	&	N/A								&	N/A		\\
		\hline
	\end{tabular}
\end{table}

From this test, we found the HoG based SVM classifier performs better than Canny based SVM classifier and is also as efficient as the CNN to detect the lane change behavior. Sometimes, it performs more stable than the CNN, although its accuracy rate is lower than that of CNNs. At the end of the detailed performance comparison shown in Fig.\ref{fig:show}, both HoG based SVM classifier and CNNs cannot detect the lane change behavior continuously, because a vehicle just overtook our vehicle, which is not present in our training data set.

\section{Conclusion}

From this project, we went through all the steps of SVM based classifier training and testing framework, and thus have handled the SVM tool for other pattern recognition applications in our future research. Moreover, we also practiced the CNN in this project as a baseline. Although the CNN is powerful, we don't know what exactly happens in the network, what the network refers to, and how to further improve it. However, after trying different features, we find a reasonable and controllable way to improve the SVM performance.

In the future, we want to integrate more information into the SVM classifier to improve the lane-change behavior detection performance, e.g. the visual odometry information from two consecutive images, the motion information from IMU sensor, and the steering information from the CANBUS. Furthermore, we can also combine many independent lane-change behavior detectors (mainly focused on the different visual features) together to implement an AdaBoost framework.


{\small
\bibliographystyle{ieee}
\bibliography{egbib}
}

\end{document}